\documentclass[11pt]{article}
\usepackage{setspace,epsfig,amsmath,amssymb,color,stmaryrd}

\usepackage{graphicx}
\usepackage{url}
\usepackage{algorithm}
\usepackage[vlined,ruled,algo2e,linesnumbered]{algorithm2e}
\usepackage{tikz}
\usepackage{pgfplots,subfigure}
\usepackage{nicefrac}       

\renewcommand{\vec}[1]{\boldsymbol{#1}}
\newcommand{\given}{\, \vert \, }
\newcommand{\svert}{\, \vert \, }
\newcommand{\cX}{\mathcal{X}}
\newcommand{\cY}{\mathcal{Y}}
\newcommand{\cH}{\mathcal{H}}
\newcommand{\cD}{\mathbf{D}}
\newcommand{\fromto}{\longrightarrow}

\begin{document}
\title{Epistemic Uncertainty Sampling\thanks{Draft version of a paper to be published in the proceedings of DS 2019, 22nd International Conference on Discovery Science, Split, Croatia, 2019.}}
 
\author{Vu-Linh Nguyen${}^a$,
S\'ebastien Destercke${}^b$,
Eyke H{\"u}llermeier${}^a$  \\[2mm]
${}^a$ Heinz Nixdorf Institute and Department of Computer Science\\
Paderborn University, Germany\\
vu.linh.nguyen@uni-paderborn.de, eyke@upb.de\\[2mm]
${}^b$ UMR CNRS 7253 Heudiasyc, Sorbonne Universit\'es,\\
Universit\'{e} de Technologie de Compi\`egne, France\\
sebastien.destercke@hds.utc.fr
}

\date{}
\maketitle              

\begin{abstract}
Various strategies for active learning have been proposed in the machine learning literature. In uncertainty sampling, which is among the most popular approaches, the active learner sequentially queries the label of those instances for which its current prediction is maximally uncertain. The predictions as well as the measures used to quantify the degree of uncertainty, such as entropy, are almost exclusively of a probabilistic nature. In this paper, we advocate a distinction between two different types of uncertainty, referred to as \emph{epistemic} and \emph{aleatoric}, in the context of active learning. Roughly speaking, these notions capture the reducible and the irreducible part of the total uncertainty in a prediction, respectively. We conjecture that, in uncertainty sampling, the usefulness of an instance is better reflected by its epistemic than by its aleatoric uncertainty. This leads us to suggest the principle of ``epistemic uncertainty sampling'', which we instantiate by means of a concrete approach for measuring epistemic and aleatoric uncertainty. In experimental studies, epistemic uncertainty sampling does indeed show promising performance.\\[2mm]
\end{abstract}

\section{Introduction}

The goal in standard supervised learning, such as binary or multi-class classification, is to learn models with high predictive accuracy from labelled training data \cite{hastie:2005,vapnik:1999}. However, labelled data does normally not come for free. On the contrary, labelling can be expensive, time-consuming, and costly. The ambition of \emph{active learning}, therefore, is to exploit labelled data in the most effective way. More specifically, the idea is to let the learning algorithm itself decide which examples it considers to be most informative. Compared to random sampling, the hope is to achieve better performance with the same amount of training data, or to reach the same performance with less data~\cite{fu:2013,settles:2010}.

The selection of training examples is often done in an iterative manner, i.e., the active learner alternates between re-training and selecting new examples. In each iteration, the usefulness of a candidate example is estimated in terms of a \textit{utility score}, and the one with the highest score is queried. In this regard, the notion of utility typically refers to uncertainty reduction: To what extent will the knowledge about the label of a specific instance help to reduce the learner's uncertainty about the sought model? In \emph{uncertainty sampling} \cite{settles:2010}, which is among the most popular approaches, utility is quantified in terms of predictive uncertainty, i.e., the active learner selects those instances for which its current prediction is maximally uncertain. The predictions as well as the measures used to quantify the degree of uncertainty, such as entropy, are almost exclusively of a probabilistic nature. Such approaches indeed proved to be successful in many applications.

Yet, as pointed out by \cite{sharma:2017}, existing approaches can be criticized for not informing about the \emph{reasons} for why an instance is considered uncertain, although this might be relevant for judging the usefulness of an example. 
In this paper, we advocate a distinction between two different types of uncertainty, referred to as \emph{epistemic} and \emph{aleatoric}\,---\,roughly speaking, these capture the reducible and the irreducible part of the total uncertainty  in a prediction, respectively. The conjecture that, in uncertainty sampling, the usefulness of an instance is better reflected by its epistemic than by its aleatoric uncertainty leads us to the idea of ``epistemic uncertainty sampling''. Our approach, which builds on a formalization of epistemic and aleatoric uncertainty as proposed by \cite{senge:2014}, is generic in the sense that is can be instantiated for any learning algorithm; concretely, we present instantiations for a Parzen window classifier, decision tree learning, and logistic regression. 

The rest of this paper is organized as follows. In the next section, we recall the general framework of uncertainty sampling and provide a brief survey of related work on active learning. In Section \ref{sec:EpisAlea}, we recall the approach of \cite{senge:2014} for modeling epistemic and aleatoric uncertainty, and then present our idea of generalizing uncertainty sampling on the basis of this approach. Instantiations of our approach for local learning (Parzen window classifier), decision tree learning and logistic regression are presented in Sections \ref{sec:Instantiations}. Experimental evaluations are given in the section \ref{sec:experiments}. The paper concludes with a short summary and an outlook on future work in Section \ref{sec:conlusion}.


\section{Uncertainty sampling}

As usual in active learning, we assume to be given a labelled set of training data $\mathbf{D}$ and a pool of unlabeled instances $\mathbf{U}$ that can be queried by the learner:
$$
\mathbf{D}=\big\{ (\vec{x}_1, y_1) , \ldots, (\vec{x}_N, y_N) \big\} , \quad 
\mathbf{U} = \big\{ \vec{x}_1, \ldots , \vec{x}_J  \big\}
$$
Instances are represented as features vectors $\vec{x}_i  = \left(x_i^1,\ldots, x_i^d \right) \in  \cX = \mathbb{R}^d$. In this paper, we only consider the case of binary classification, where labels $y_i$ are taken from $ \cY = \lbrace 0, 1 \}$, leaving the more general case of multi-class classification for future work. We denote by $\cH \subset \cY^\cX$ the underlying hypothesis space, i.e., the class of candidate models $h:\, \cX \fromto \cY$ the learner can choose from. Often, hypotheses are parametrized by a parameter vector $\theta \in \Theta$; in this case, we equate a hypothesis $h= h_\theta \in \cH$ with the parameter $\theta$, and the model space $\cH$ with the parameter space $\Theta$.

In uncertainty sampling, instances are queried in a greedy fashion. 
Given the current model $\theta$ that has been trained on $\mathbf{D}$, each instance $\vec{x}_j$ in the current pool $\mathbf{U}$ is assigned a \textit{utility} score $s(\theta,\vec{x}_j)$, and the next instance to be queried is the one with the highest score \cite{lewis:1994,settles:2010,sharma:2017}. The chosen instance is labelled (by an oracle or expert) and added to the training data $\mathbf{D}$, on which the model is then re-trained. The active learning process for a given budget $B$ (i.e, the number of unlabelled instances to be queried) is summarized in Algorithm \ref{alg:genracalgo}.

\begin{algorithm2e}
	\KwIn{$\mathbf{U}$, $\mathbf{D}$, $\theta$- initial pool, training data, classifier, and $B$-budget}
	\KwOut{$\mathbf{U}$, $\mathbf{D}$, $\theta$ - updated pool, training data, classifier}
	initialize $b = 0$\;
	\While{ $b < B$} {
		\ForEach{ $\vec{x}\in \mathbf{U}$}{compute $s(\theta,\vec{x})$}
	query the label of the optimal instance $\vec{x}^*$ with respect to $s(\theta,\vec{x})$
	$\mathbf{D} = \mathbf{D} \cup \{\vec{x}^*, y^*  \} $ \;
	$\mathbf{U} = \mathbf{U} \setminus \{\vec{x}^*, y^* \} $ \;
	train $\theta$ from $\mathbf{D}$\;
	$b = b +1$\;
	}
	\textbf{Return} $\mathbf{U}$, $\mathbf{D}$, $\theta$\;
	\caption{Uncertainty sampling}
	\label{alg:genracalgo}
\end{algorithm2e}  

Assuming a probabilistic model producing predictions in the form of probability distributions $p_\theta( \cdot \given \vec{x})$ on $\cY$, the utility score is typically defined in terms of a measure of uncertainty. Thus, instances on which the current model is highly uncertain are supposed to be maximally informative \cite{settles:2010,sharma:2017}. Popular examples of such measures include   
\begin{itemize}
\item[--] the entropy:
\begin{align}
s(\theta,\vec{x}) =  - \sum_{\lambda \in \cY} p_{\theta}(\lambda \given\vec{x})\log p_{\theta}(\lambda \given\vec{x}) \, ,
\end{align} 
\item[--] the least confidence:
\begin{align}
s(\theta,\vec{x}) = 1 - \max_{\lambda \in \cY} p_{\theta}(\lambda \given\vec{x}) \, ,\label{eq:standardsampling}
\end{align}
\item[--] the smallest margin:
\begin{align}
s(\theta,\vec{x})=  p_{\theta}(\lambda_n\given \vec{x}) - p_{\theta} &(\lambda_m\given \vec{x}) \, , 
\end{align}
where 
$\lambda_m = \arg\max_{\lambda\in \cY}  p_{\theta}(\lambda\given \vec{x})$ and $\lambda_n = \arg\max_{\lambda \in \cY \setminus \lambda_m}  p_{\theta}(\lambda \given \vec{x})$.
\end{itemize}
All the three measures ought to be maximized. In the case of binary classification, i.e, $\cY = \{ 0, 1 \}$, all these measures rank unlabelled instances in the same order and look for instances with small difference between $ p_{\theta}(0 \given \vec{x})$ and  $p_{\theta}(1 \given \vec{x})$.  

\section{Epistemic and aleatoric uncertainty}\label{sec:EpisAlea}

A main building block of our approach to active learning is the distinction between the \emph{epistemic} and \emph{aleatoric} uncertainty involved in the prediction for an instance $\vec{x}$. Although this distinction is well accepted in the literature on uncertainty \cite{hora:1996}, it has been considered in machine learning only very recently \cite{kend_wu:17,nguyen:2018,senge:2014}. Here, we adopt the formal model proposed by \cite{senge:2014}, which is based on the use of relative likelihoods, historically proposed by \cite{birnbaum:1962} and then justified in other settings such as possibility theory~\cite{walley:1999}. For the sake of completeness and self-containedness, we briefly recall the essence of this approach.

As before, we proceed from an instance space $\cX$, an output space $\cY = \{ 0, 1 \}$ encoding the two classes, and a hypothesis space $\mathcal{H}$ consisting of probabilistic classifiers $h: \cX \longrightarrow [0,1]$. We denote by $p_{h}(1 \svert \vec{x}) = h(\vec{x})$ and $p_{h}(0 \svert \vec{x}) = 1- h(\vec{x})$ the (predicted) probability that instance $\vec{x} \in \cX$ belongs to the positive and negative class, respectively. Given a set of training data $\cD = \{ (\vec{x}_i , y_i) \}_{i=1}^N \subset \cX \times \cY$, the normalized likelihood of a model $h$ is defined as
\begin{align}\label{eq:plausibility}
\pi_{\mathcal{H}}(h) = \frac{L(h)}{L(h^{ml})} =  
\frac{L(h)}{\max_{h' \in \mathcal{H}} L(h')} \enspace ,
\end{align}  
where $L(h) = \prod_{i=1}^N p_{h}(y_i \svert \vec{x}_i)$ is the likelihood of $h$, and $h^{ml} \in \mathcal{H}$ the maximum likelihood estimation on the training data.
For a given instance $\vec{x}$, the degrees of support (plausibility) of the two classes are defined as follows: 
\begin{eqnarray}
\pi(1\svert \vec{x}) & =  & \sup_{h \in \mathcal{H}} \min \big[\pi_{\mathcal{H}}(h), p_{h}(1 \svert \vec{x}) - p_{h}(0 \svert \vec{x}) \big], \label{eq:support posx}\\
\pi(0 \svert \vec{x}) & = & \sup_{h \in \mathcal{H}} \min \big[\pi_{\mathcal{H}}(h), p_{h}(0 \svert \vec{x}) - p_{h}(1 \svert \vec{x}) \big]. \label{eq:support negx}
\end{eqnarray} 
So, $\pi(1 \svert  \vec{x})$ is high if and only if a highly plausible model supports the positive class much stronger (in terms of the assigned probability mass) than the negative class (and $\pi(0 \svert \vec{x})$ can be interpreted analogously)\footnote{Technically, we assume that, for each $\vec{x} \in \mathcal{X}$, there are hypotheses $h,h' \in \mathcal{H}$ such that $h(\vec{x}) \geq 0.5$ and $h'(\vec{x}) \leq 0.5$, which implies $\pi(1\svert \vec{x}) \geq 0$ and $\pi(0 \svert \vec{x}) \geq 0$.}.
Note that, with $f(a)= 2a-1$, we can also rewrite \eqref{eq:support posx}--\eqref{eq:support negx} as follows:
\begin{eqnarray}
\pi(1\svert \vec{x}) & =  & \sup_{h \in \mathcal{H}} \min \big[\pi_{\mathcal{H}}(h), f(h(\vec{x})) \big], \label{eq:support pos}\\
\pi(0 \svert \vec{x}) & = & \sup_{h \in \mathcal{H}} \min \big[\pi_{\mathcal{H}}(h), f(1- h(\vec{x})) \big]. \label{eq:support neg}
\end{eqnarray} 
Given the above degrees of support, the degrees of epistemic uncertainty $u_e$ and aleatoric uncertainty $u_a$ are defined as follows:
\begin{eqnarray}
u_e(\vec{x}) & = & \min \big[ \pi(1  \svert \vec{x}), \pi(0 \svert \vec{x}) \big] \, , \label{eq:epistemic} \\
u_a(\vec{x}) & = & 1 - \max \big[ \pi(1  \svert \vec{x}), \pi(0  \svert \vec{x}) \big] \, .\label{eq:aleatoric}
\end{eqnarray}  
Thus, epistemic uncertainty refers to the case where both the positive and the negative class appear to be plausible, while the degree of aleatoric uncertainty \eqref{eq:aleatoric} is the degree to which none of the classes is supported. These uncertainty degrees are completed with degrees $s_{1}(\vec{x})$ and $s_{0}(\vec{x})$ of (strict) preference in favor of the positive and negative class, respectively:
$$
s_{1}(\vec{x}) = \left\{ 
\begin{array}{cl}
1 - (u_a(\vec{x})+ u_e(\vec{x})) & \text{ if } \pi(1 \svert \vec{x}) >\pi(0 \svert  \vec{x}), \\[2mm]
\frac{1 - (u_a(\vec{x})+ u_e(\vec{x}))}{2} & \text{ if } \pi(1\svert \vec{x}) = \pi(0 \svert \vec{x}), \\[2mm] 
0 & \text{ if } \pi(1\svert \vec{x}) < \pi(0 \svert \vec{x}).
\end{array} \right.
$$
With an analogous definition for $s_{0}(\vec{x})$, we have 
$s_{0}(\vec{x}) + s_{1}(\vec{x})+ u_a(\vec{x})+ u_e(\vec{x}) \equiv 1$. 
Besides, it has the following properties: 
\begin{itemize}
\item[-] $s_{1}(\vec{x})$ ($s_{0}(\vec{x})$) will be high if and only if, for all plausible models, the probability of the positive (negative) class is significantly higher than the one of the negative (positive) class; 
\item[-] $u_e(\vec{x})$ will be high if class probabilities strongly vary within the set of plausible models, i.e., if we are unsure how to compare these probabilities. In particular, it will be $1$ if and only if we have $h(\vec{x})=1$ and $h'(\vec{x})=0$ for two totally plausible models $h$ and $h'$; 
\item[-] $u_a(\vec{x})$ will be high if class probabilities are similar for all plausible models, i.e., if there is strong evidence that $h(\vec{x}) \approx 0.5$. In particular, it will be close to $1$ if all plausible models allocate their probability mass around $h(\vec{x})=0.5$.
\end{itemize}    
Roughly speaking, aleatoric uncertainty is due to influences on the data-generating process that are inherently random, whereas epistemic uncertainty is caused by a lack of knowledge. Or, stated differently, $u_e$ and $u_a$ measure the \textit{reducible} and the \textit{irreducible} part of the total uncertainty, respectively. It thus appears reasonable to assume that epistemic uncertainty is more relevant for active learning: While it makes sense to query additional class labels in regions where uncertainty can be reduced, doing so in regions of high aleatoric uncertainty appears to be less reasonable. This leads us to the principle of \emph{epistemic uncertainty sampling}, which prescribes the selection 
\begin{align}
\vec{x}^*  = &\arg\max_{\vec{x} \in \mathbf{U}} u_e(\vec{x}) \, .
\label{eq:epist}
\end{align}
For comparison, we will also consider an analogous selection rule based on the aleatoric uncertainty, i.e., 
\begin{align}
\vec{x}^*  = &\arg\max_{\vec{x} \in \mathbf{U}} u_a(\vec{x}) \, .
\label{eq:alea}
\end{align}

Let us note that the above approach is completely generic and can in principle be instantiated with any hypothesis space $\mathcal{H}$. The uncertainty measures (\ref{eq:epist}--\ref{eq:alea}) can be derived very easily from the support degrees (\ref{eq:support pos}--\ref{eq:support neg}). 
The computation of the latter may become difficult, however, as it requires the solution of an optimization problem, the properties of which depend on the choice of $\mathcal{H}$. 

\section{Instantiations of the general approach}
\label{sec:Instantiations}
We are going to present practical methods to determine (\ref{eq:support pos}--\ref{eq:support neg}) for the cases of local learning and logistic regression in Sections \ref{sec:ApLocal} and \ref{sec:ApLogist}, respectively. 

\subsection{Local learning}
\label{sec:ApLocal}

This section presents an instantiation of our approach for the case of local learning using a Parzen window classifier \cite{chapelle:2005}. The method is then adapted to the case where the decision tree classifier \cite{quinlan:1986,safavian:1991} is employed as the based learner.

As already said, instantiating the approach essentially means to address the question of how to compute the degrees of support (\ref{eq:support pos}--\ref{eq:support neg}), from which everything else can easily be derived. 

By local learning, we refer to a class of non-parametric models that derive predictions from the training information in a local region of the instance space, for example the local neighborhood of a query instance \cite{bottou:1992,cover:1967}. As a simple example, we consider the Parzen window classifier \cite{chapelle:2005}, to which our approach can be applied in a quite straightforward way.  
To this end, for a given instance $\vec{x}$, define the set of its neighbours as follows:
\begin{align}
R(\vec{x},\epsilon) = \big\{ (\vec{x}_i , y_i) \in \mathbf{D} \given \| \vec{x}_i  - \vec{x} \| \leq \epsilon \big\} \, ,
\end{align}
where  
$\epsilon$ is the width of the Parzen window (a practical method to determine such a width will be given latter).   

In binary classification, a local region $R$ can be associated with a constant hypothesis $h_\theta$, $\theta \in \Theta = [0,1]$, where $h_\theta(\vec{x}) \equiv \theta$ is the probability of the positive class in the region; thus, $h_\theta$ predicts the same probabilities $p_h(1 \given \vec{x}) = \theta$ and $p_h( 0 \given \vec{x}) = 1- \theta$ for all $\vec{x} \in R$. The underlying hypothesis space is given by $\mathcal{H} = \{ h_\theta \given 0 \leq \theta \leq 1 \}$.
With $n$ and $p$ the number of positive and negative instances, respectively, within a Parzen window $R(\vec{x}, \epsilon)$, the likelihood and the maximum likelihood estimate of $\theta$ are respectively given by 
\begin{align}
L(\theta)=
\left( \begin{array}{c}
n+p \\
n \\
\end{array} \right) 
\theta^n (1-\theta)^p  \, 
\text{ and }
\hat{\theta} =\frac{n}{n+p} \, . 
\end{align}
Therefore, the degrees of support for the positive and negative classes are 
\begin{align}
\pi(1\given \vec{x}) = \sup_{\theta \in [0,1]} &\min \left( \frac{\theta^p(1 - \theta)^n}{\big(\frac{p}{n+p}\big)^p \big(\frac{n}{n+p}\big)^n} , \, 2\theta-1 \right)  \, , \label{eq:knn sup pos}\\[2mm]
\pi(0\given \vec{x}) = \sup_{\theta \in [0,1]} &\min \left( \frac{\theta^p(1 - \theta)^n}{\big(\frac{p}{n+p}\big)^p \big(\frac{n}{n+p}\big)^n}, \, 1-2\theta \right)  \, . \label{eq:knn sup neg} 
\end{align}

\begin{figure}
\begin{center}
\includegraphics[scale=0.25]{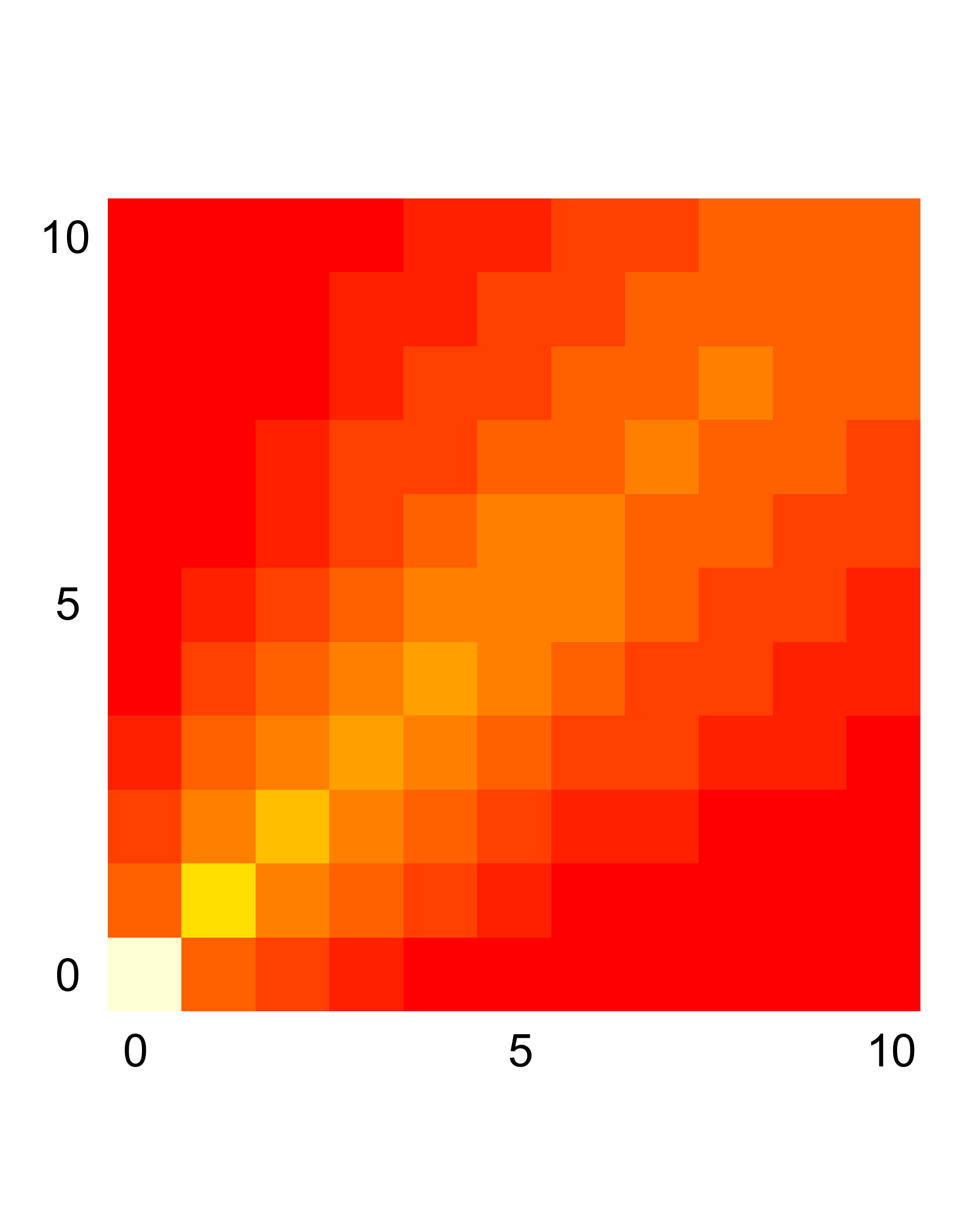}
\includegraphics[scale=0.25]{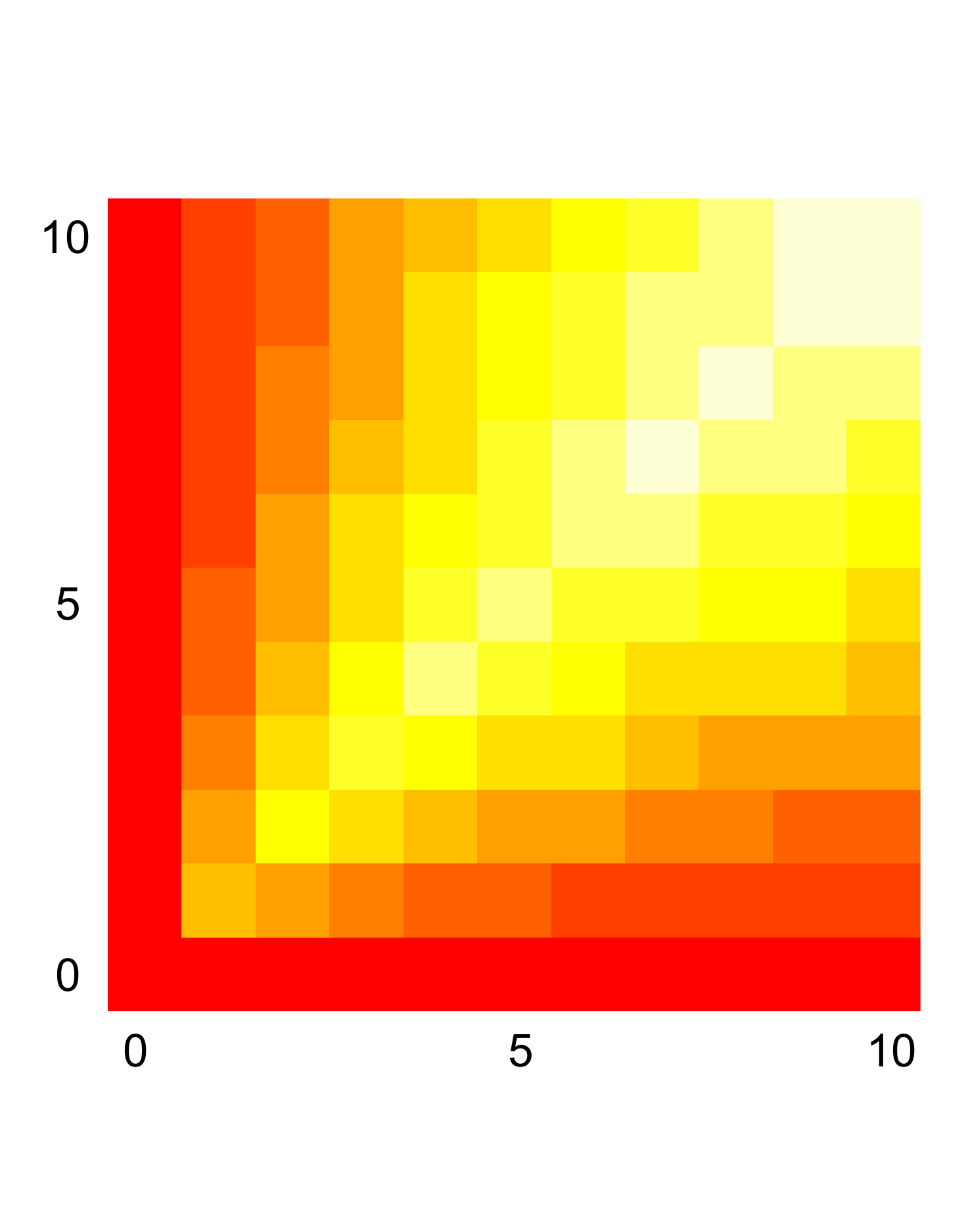}
\includegraphics[scale=0.25]{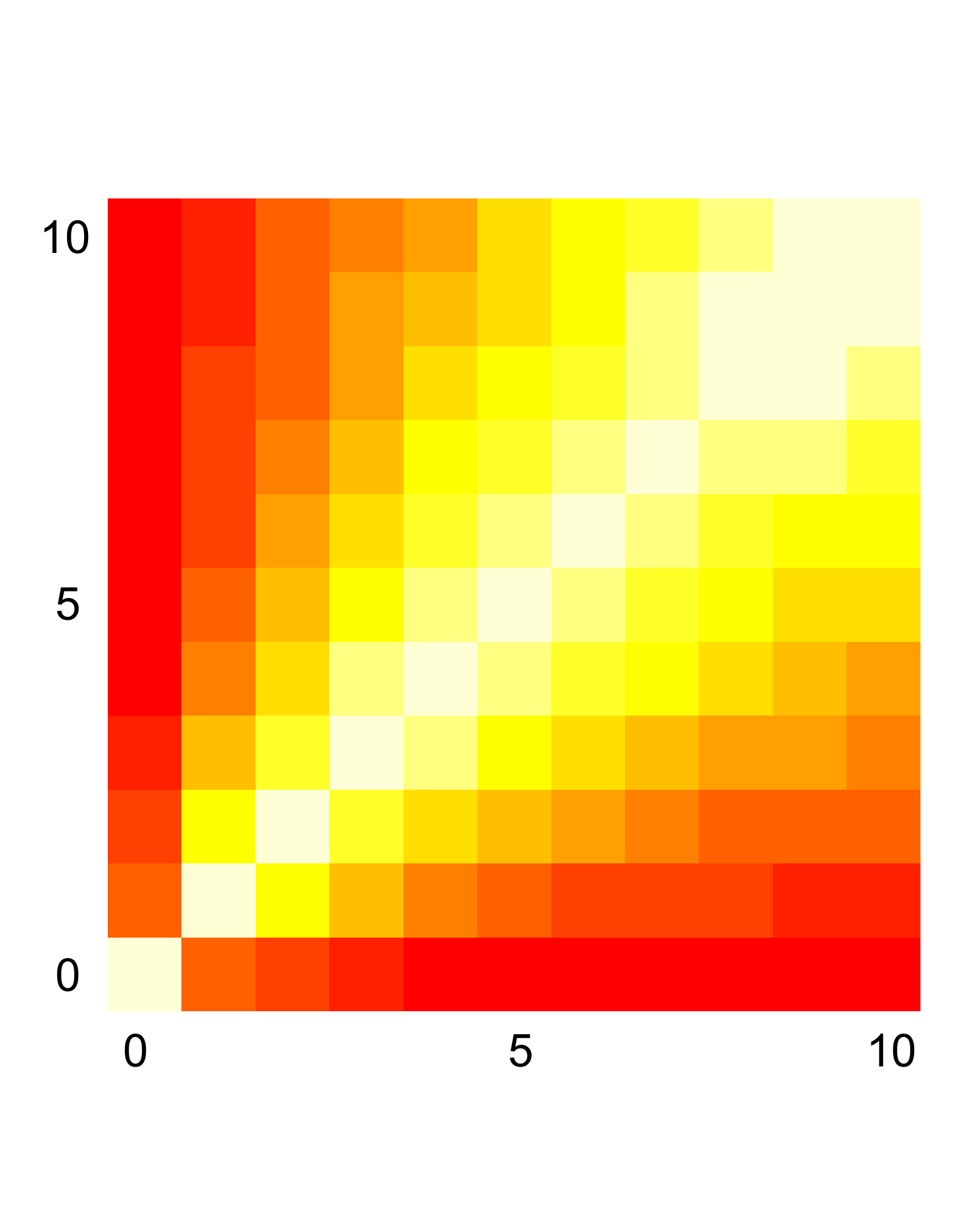} 
\vspace*{-8mm}
\caption{From left to right: Epistemic, aleatoric, and total uncertainty (epistemic $+$ aleatoric) as a function of the numbers $p, n \in \{0,1, \ldots , 10 \}$ of positive and negative examples in a region (Parzen window) of the instance space (lighter colors indicate higher values).}
\label{fig:heat}
\end{center}
\end{figure}

Solving \eqref{eq:knn sup pos} and \eqref{eq:knn sup neg} comes down to maximizing a scalar function over a bounded domain, for which standard solvers can be used. We applied Brent's method\footnote{For an implementation in Python, see \url{https://docs.scipy.org/doc/scipy-0.19.1/reference/generated/scipy.optimize.minimize_scalar.html}} (which is a variant of the golden section method) to find a local minimum in the interval $\theta \in [0,1]$. 
From (\ref{eq:knn sup pos}--\ref{eq:knn sup neg}), the epistemic and aleatoric uncertainty associated with the region $R$ can be derived according to (\ref{eq:epist}) and (\ref{eq:alea}), respectively. For different combinations of $n$ and $p$, these uncertainty degrees can be pre-computed (cf.\ Figure \ref{fig:heat}).

How to determine the width $\epsilon$ of the Parzen window? This value is difficult to assess, and an appropriate choice strongly depends properties of the data and the dimensionality of the instance space. Intuitively, it is even difficult to say in which range this value should lie. Therefore, instead of fixing $\epsilon$, we fixed an absolute number $K$ of neighbors in the training data, which is intuitively more meaningful and easier to interpret. A corresponding value of $\epsilon$ is then determined in such a way that the average number of nearest neighbours of  instances $\vec{x}_i$ in the training data $\mathbf{D}$ is just $K$ (see Algorithm \ref{alg:determineEps}). In other words, $\epsilon$ is determined indirectly via $K$. 

Since $K$ is an average, individual instances may have more or less neighbors in their Parzen windows. In particular, a Parzen window may also be empty. In this case, we set $u_e(\vec{x})=1$ by definition, i.e., we consider this as a case of full epistemic uncertainty. Likewise, the uncertainty is considered to be maximal for all other sampling techniques. If the accuracy of the Parzen classifier needs to be determined, we assume that it yields a wrong prediction.  

\begin{algorithm2e}
	\KwIn{$\mathbf{D}$-normalized data, $K$-number}
	\KwOut{the local width $\epsilon_{K}$}
	\ForEach{ $\vec{x}_n \in \mathbf{D}$}{
	\ForEach{ $\vec{x}_m \neq \vec{x}_n$}{compute $d\big(\vec{x}_n, \vec{x}_m\big)$\;}
	form $1 \times (n-1)$ vector $\mathbf{d}_n = \big(d\big(\vec{x}_n, \vec{x}_m\big)\given n\neq m \big)$\;
	sort $\mathbf{d}_n$ by increasing order and determine the $K$-th element $\mathbf{d}_n^K$\;}
	\Return $\epsilon_{K} = \frac{\sum_{n=1}^{\given \mathbf{D} \given}\mathbf{d}_n^K}{\given \mathbf{D} \given}$\; 
	\caption{Determining the width $\epsilon$.}
	\label{alg:determineEps}
\end{algorithm2e}  

In a similar way, the approach can be applied to decision tree learning \cite{quinlan:1986,safavian:1991}. In fact recall that a decision tree partitions the instance space $\mathcal{X}$ into (rectangular) regions $R_1, \ldots , R_L$ (i.e., $\bigcup_{i=1}^L R_i = \mathcal{X}$ and $R_i \cap R_j = \emptyset$ for $i \neq j$) associated with corresponding leafs of the tree (each leaf node defines a region $R$). Again, in the case of binary classification, we can assume each region $R$ to be associated with a constant hypothesis $h_\theta$, $\theta \in \Theta = [0,1]$, where $h_\theta(\vec{x}) \equiv \theta$ is the probability of the positive class. Therefore, degrees of epistemic and aleatoric uncertainty degrees can be derived in the same way as described above.

\subsection{Logistic regression}
\label{sec:ApLogist}

In this section, we present another instantiation of our approach for a commonly used learning algorithm, namely logistic regression. In contrast to nonparametric, local learning methods such as the Parzen window classifier, logistic regression is a parametric class of linear models, and hence coming with comparatively restrictive assumptions.  



Recall that logistic regression assumes posterior probabilities to depend on feature vectors $\vec{x}  = (x^1,\ldots, x^d) \in \mathbb{R}^d$ in the following way:
\begin{align}
h(\vec{x}) = p(1 \svert \vec{x})= \frac{\exp \left( \theta_0 + \sum_{i= 1}^d \theta_i \, x^i \right)}{1 + \exp \left( \theta_0 + \sum_{i = 1}^d \theta_i \, x^i \right) }  
\end{align} 
This means that learning the model comes down to estimating a parameter vector $\theta =(\theta_0, \ldots, \theta_d)$, which is commonly done through likelihood maximization \cite{menard:2002}.         
To avoid numerical issues (e.g, having to deal with the exponential function for large $\theta$) when maximizing the target function, we employ $L_2$-regularization. The corresponding version of the log-likelihood function \eqref{eq:logLikelihoodReg} is strictly concave \cite{rennie:2005}:   
\begin{align}\label{eq:logLikelihoodReg}
l(\theta) = \log L(\theta) &= \sum_{n=1}^N  y_n \left(  \theta_0 + \sum_{i=1}^d \theta_i x_n^i \right) \\ 
&- \sum_{n=1}^N \ln \left(1+ \exp \left( \theta_0 + \sum_{i=1}^d \theta_i x_n^i \right) \right) - \frac{\gamma}{2}\sum_{i=0}^d \theta_i^2 \nonumber,
\end{align}   
where the regularization term $\gamma$ will be fixed to $1$. 

We now focus on determining the degree of support  \eqref{eq:support pos} for the positive class, and then summarize the results for the negative class (which can be determined in a similar manner). Associating each hypothesis $h \in \mathcal{H}$ with a vector $\theta \in \mathbb{R}^{d+1}$, the degree of support \eqref{eq:support pos} can be rewritten as follows: 
\begin{align}
\pi(1\svert \vec{x}) =  \sup_{\theta \in \mathbb{R}^{d+1}} \min \big[\pi(\theta), 2h(\vec{x})-1 \big]  \label{eq:support pos re}
\end{align}
It is easy to see that the target function to be maximized in \eqref{eq:support pos re} is not necessarily concave.
Therefore, we propose the following approach. 

Let us first note that whenever $h(\vec{x}) < 0.5$, we have $2h(\vec{x})-1 \leq 0$ and $\min \big[\pi_{\mathcal{H}}(h), 2h(\vec{x})-1 \big] \leq 0$. Thus the optimal value of the target function \eqref{eq:support pos} can only be achieved for some hypotheses $h$ such that $h(\vec{x}) \in [0.5,1]$. 
For a given value $\alpha \in [0.5,1]$, the set of hypotheses $h$ such that $h(\vec{x}) = \alpha$ corresponds to the convex set 
\begin{align} \label{eq:invIm}
\theta^{\alpha} = \bigg\{ \theta \,\big\vert\, \theta_0 + \sum_{i=1}^d \theta_i x^i  = \ln\bigg(\frac{\alpha}{1-\alpha}\bigg) \bigg \}  \, . 
\end{align} 
The optimal value $\pi_{\alpha}^*(1\svert \vec{x})$ that can be achieved within the region \eqref{eq:invIm} can be determined as follows:
\begin{align}\label{eq:localSup}
\pi_{\alpha}^*(1\svert \vec{x})&= \sup_{\theta \in \theta^{\alpha}} \min \big[\pi(\theta),2\alpha-1\big] = \min \big[\sup_{\theta \in \theta^{\alpha}}\pi(\theta), 2\alpha-1 \big] \,.
\end{align}  
Thus, to find this value, we maximize the concave log-likelihood over a convex~set:
\begin{align}\label{eq:optiPro}
\theta^{*}_{\alpha} = \arg\sup_{\theta \in \theta^{\alpha}}l(\theta)
\end{align}
As the log-likelihood function \eqref{eq:logLikelihoodReg} is concave and has second-order derivatives, we tackle the problem with a Newton-CG algorithm \cite{nocedal:2006}. Furthermore, the optimization problem \eqref{eq:optiPro} can be solved using sequential least squares programming\footnote{For an implementation in Python, see \url{https://docs.scipy.org/doc/scipy/reference/generated/scipy.optimize.minimize.html}} \cite{philip:2010}. Since regions defined in \eqref{eq:invIm} are parallel hyperplanes, the solution of the optimization problem \eqref{eq:support pos} can then be obatined by solving the following problem:
\begin{align} \label{eq:supPoEst}
\sup_{\alpha \in [0.5,1)} \pi^*_{\alpha}(1\vert \vec{x}) 
= \sup_{\alpha \in [0.5,1)} \min \big[\pi(\theta^{*}_{\alpha}), 2\alpha-1 \big] \,.
\end{align} 
Following a similar procedure, we can estimate the degree of support for the negative class \eqref{eq:support neg} as follows:
\begin{align} \label{eq:supNeEst}
 \sup_{\alpha \in (0,0.5]}\pi_{\alpha}^*(0\vert \vec{x})= \sup_{\alpha \in (0,0.5]} \min \big[\pi(\theta^{*}_{\alpha}), 1- 2\alpha\big]  
\end{align} 
Note that limit cases $\alpha = 1$ and $\alpha = 0$ cannot be solved, since the region \eqref{eq:invIm} is then not well-defined (as $\ln(\infty)$ and $\ln(0)$ do not exist). For the purpose of practical implementation, we handle \eqref{eq:supPoEst} by discretizing the interval over $\alpha$. That is, we optimize the target function for a given number of values $\alpha \in [0.5,1)$ and consider the solution corresponding to the $\alpha$ with the highest optimal value of the target function $\pi_{\alpha}^*(1\svert \vec{x})$ as the maximum estimator. Similarly, \eqref{eq:supNeEst} can be handled over the domain $(0,0.5]$. 

In practice, we evaluate \eqref{eq:supPoEst} and \eqref{eq:supNeEst} on uniform discretizations of cardinality $50$
of $[0.5,1)$ and $(0,0.5]$, respectively. We can further increase efficiency by avoiding computations for values of $\alpha$ for which we know that $2 \alpha -1$ and $1-2\alpha$ are lower than the current highest support value given to class $1$ and $0$, respectively. See Algorithm \ref{alg:EpisAlea} for a pseudo-code description of the whole procedure.

\begin{algorithm2e}
	\KwIn{$Q$, $\mathbf{D}$, $\theta^{ml}$, $\vec{x}$- initial pool, training data, classifier, unlabelled instance}
	\KwOut{$\pi(1 \svert \vec{x})$, $\pi(0 \svert \vec{x})$ - degrees of support}
	initialize subsets $Q_p$, $Q_n$ of cardinality $Q$\;
	$\pi(1 \svert \vec{x}) = \max(2h^{ml}(\vec{x})-1,0)$ , $\pi(0 \svert \vec{x}) = \max(1-2h^{ml}(\vec{x}),0)$ \;
	\For{$q = 1, \ldots, Q$}{
    $\alpha_p = \max(Q_p)$; $\alpha_n = \min(Q_n)$ \;
	\If{$2\alpha_p -1 > \pi(1 \svert \vec{x})$}{solve \eqref{eq:optiPro} for $\vec{x}$, $\alpha_p$ and return $\theta$\;
	$\pi(1 \svert \vec{x}) = \max(\pi(1 \svert \vec{x}), \min(\pi_{\mathcal{H}}(\theta),2\alpha_p -1))$ \;}
	\If{$1 - 2\alpha_n > \pi(0 \svert \vec{x})$}{solve \eqref{eq:optiPro} for $\vec{x}$, $\alpha_n$ and return $\theta$\;
	$\pi(0 \svert \vec{x}) = \max(\pi(0 \svert \vec{x}), \min(\pi_{\mathcal{H}}(\theta),1- 2\alpha_p))$ \;}
	$Q_p = Q_p \setminus \{\alpha_p\}$, $Q_n = Q_n \setminus \{\alpha_n\}$ \;
	}
	\textbf{Return} $\pi(1 \svert \vec{x})$, $\pi(0 \svert \vec{x})$ \;
	\caption{Degrees of support for logistic regression}
	\label{alg:EpisAlea}
\end{algorithm2e} 

\section{Experimental results}
\label{sec:experiments}



To illustrate the performance of our uncertainty measures in active learning, we conducted experiments on data sets from the UCI repository\footnote{\url{http://archive.ics.uci.edu/ml/index.php}}, the main properties of which are summarized in Table \ref{tab:datasets}. 

\begin{table}
\centering
\caption{Data sets used in the experiments}\label{tab:datasets}
\begin{tabular}{ccccc}
\hline
\# &name &  \# instances & \# features & attributes \\
\hline
1 &parkinsons               &197  &22  &real\\
2 &vertebral-column         &310  &6   &real\\
3 &ionosphere               &351  &34  &real\\ 
4 &climate-model            &540  &18  &real\\
5 &breast-cancer            &569  &30  &real\\
6 &blood-transfusion        &748  &5   &real\\
7 &QSAR                     &1055 &41  &integer, real \\
8 &banknote-authentication  &1372 &4   &real\\
\hline
\end{tabular}
\end{table}

 



\subsection{Local learning}\label{sec:exLocal}

We follow a $10$-fold cross-validation procedure, considering each fold as the test set, while the other folds are used for learning. The latter is randomly split into a training data set and a pool set. The proportions of training/pool/test sets are $10/80/10\, \%$ and accuracies are averaged. The budget of the active learner is fixed to be $30\%$ of the original data.

\input{FigLocalAcc}

After each query, we update the data sets and, correspondingly, the classifiers. The improvements of the classifiers are compared for four different uncertainty measures, i.e., uncertainty sampling (following the strategy presented in Algorithm \ref{alg:genracalgo}) based on four measures for selecting unlabelled instances: random sampling, standard uncertainty \eqref{eq:standardsampling}, epistemic uncertainty \eqref{eq:epistemic}, aleatoric uncertainty \eqref{eq:aleatoric}.

To reduce the computational efforts, in each iteration, the learner is allowed to evaluate and query instances from a randomly selected subset consisting of $10\%$ of the data in the pool. Since we are not, in the first place, interested in maximizing performance, but in analyzing the effectiveness of active learning approaches, we simply fix the neighborhood size $K$ as the square root of the size of the data set (number of instances in the initial training set and pool) \cite{lall:1996}.

As can be seen in Figure \ref{fig:UncAcc8}, the results are nicely in agreement with our expectations: Epistemic uncertainty sampling performs the best and aleatoric uncertainty sampling the worst. Moreover, standard uncertainty sampling and random sampling are in-between the two. This supports our conjecture that, from an active learning point of view, epistemic uncertainty is the more useful information. Even if the improvements compared to standard uncertainty sampling are not huge, they are still visible and quite consistent.

The results for decision tree learning (cf.\ Figure \ref{fig:UncAccDecisiontree}) are quite similar and again in agreement with our expectations.




\input{FigTreeAcc}

\subsection{Logistic regression}
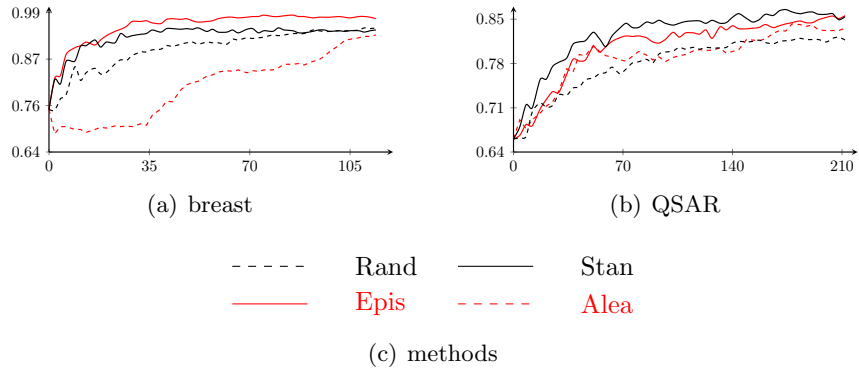
\begin{figure}[!ht]
    \centering    
        \subfigure[breast]{
        \begin{tikzpicture}[remember picture,scale=0.55]
            \begin{axis}[
                    axis lines=left,
                    xmin=0,
                    xmax=120,
                    ytick={.64,.755,...,1},
                    xtick={0,35,...,115},
                    ymin=.64,
                    ymax=1,
                    thick,
                    width=0.66\textwidth,
                    height=0.34\textwidth
                ]
                \addplot[red,smooth,thick] plot coordinates {
                  (0, 0.7455)	(2, 0.8226)	(4, 0.8295)	(6, 0.882)	(9, 0.8981)	(11, 0.9033)	(13, 0.9121)	(16, 0.9049)	(18, 0.9173)	(20, 0.9279)	(23, 0.9385)	(25, 0.9438)	(27, 0.9508)	(29, 0.9631)	(32, 0.9631)	(34, 0.9649)	(36, 0.9596)	(39, 0.9579)	(41, 0.9684)	(43, 0.9701)	(46, 0.9684)	(48, 0.9701)	(50, 0.9649)	(52, 0.9614)	(55, 0.9631)	(57, 0.9649)	(59, 0.9701)	(62, 0.9737)	(64, 0.9737)	(66, 0.9719)	(69, 0.9737)	(71, 0.9754)	(73, 0.9754)	(75, 0.9789)	(78, 0.9754)	(80, 0.9754)	(82, 0.9772)	(85, 0.9789)	(87, 0.9754)	(89, 0.9754)	(92, 0.9719)	(94, 0.9737)	(96, 0.9737)	(98, 0.9737)	(101, 0.9754)	(103, 0.9737)	(105, 0.9719)	(108, 0.9754)	(110, 0.9754)	(112, 0.9737)	(114, 0.9701)};
                 \addplot[red,dashed,thick] plot coordinates {
                  (0, 0.7455)	(2, 0.6853)	(4, 0.7029)	(6, 0.7029)	(9, 0.6958)	(11, 0.6958)	(13, 0.6888)	(16, 0.6958)	(18, 0.7011)	(20, 0.6993)	(23, 0.7046)	(25, 0.7029)	(27, 0.7046)	(29, 0.7029)	(32, 0.7099)	(34, 0.7064)	(36, 0.7309)	(39, 0.7467)	(41, 0.766)	(43, 0.7643)	(46, 0.7941)	(48, 0.8064)	(50, 0.8116)	(52, 0.8151)	(55, 0.8274)	(57, 0.8309)	(59, 0.8379)	(62, 0.8414)	(64, 0.8397)	(66, 0.8432)	(69, 0.8362)	(71, 0.8414)	(73, 0.8397)	(75, 0.8432)	(78, 0.8432)	(80, 0.8485)	(82, 0.8572)	(85, 0.8572)	(87, 0.859)	(89, 0.8555)	(92, 0.8643)	(94, 0.8679)	(96, 0.8768)	(98, 0.8893)	(101, 0.9034)	(103, 0.9157)	(105, 0.9192)	(108, 0.9227)	(110, 0.9262)	(112, 0.9262)	(114, 0.9297)};
                  \addplot[smooth,thick,dashed] plot coordinates {
                (0, 0.7455)	(2, 0.7433)	(4, 0.7717)	(6, 0.7821)	(9, 0.8507)	(11, 0.8174)	(13, 0.8367)	(16, 0.8471)	(18, 0.8331)	(20, 0.8367)	(23, 0.8578)	(25, 0.8682)	(27, 0.8665)	(29, 0.8787)	(32, 0.8875)	(34, 0.8876)	(36, 0.8929)	(39, 0.8946)	(41, 0.9034)	(43, 0.9017)	(46, 0.9069)	(48, 0.9104)	(50, 0.9122)	(52, 0.9069)	(55, 0.9122)	(57, 0.9104)	(59, 0.9104)	(62, 0.9069)	(64, 0.9192)	(66, 0.9174)	(69, 0.9227)	(71, 0.921)	(73, 0.9227)	(75, 0.9245)	(78, 0.9298)	(80, 0.928)	(82, 0.9297)	(85, 0.9332)	(87, 0.9332)	(89, 0.9333)	(92, 0.942)	(94, 0.9403)	(96, 0.9385)	(98, 0.9385)	(101, 0.9403)	(103, 0.9403)	(105, 0.9403)	(108, 0.942)	(110, 0.9473)	(112, 0.9456)	(114, 0.9473)};
                  \addplot[smooth,thick,thick] plot coordinates {
                  (0, 0.7455)	(2, 0.8192)	(4, 0.8087)	(6, 0.8647)	(9, 0.8649)	(11, 0.9017)	(13, 0.9034)	(16, 0.9174)	(18, 0.8998)	(20, 0.9122)	(23, 0.9103)	(25, 0.9333)	(27, 0.9245)	(29, 0.928)	(32, 0.9315)	(34, 0.9333)	(36, 0.942)	(39, 0.9368)	(41, 0.9456)	(43, 0.9456)	(46, 0.9438)	(48, 0.9456)	(50, 0.9438)	(52, 0.9473)	(55, 0.9385)	(57, 0.9438)	(59, 0.9368)	(62, 0.9368)	(64, 0.942)	(66, 0.935)	(69, 0.9438)	(71, 0.9456)	(73, 0.9351)	(75, 0.9403)	(78, 0.9386)	(80, 0.9473)	(82, 0.9421)	(85, 0.9403)	(87, 0.9351)	(89, 0.9351)	(92, 0.9386)	(94, 0.9421)	(96, 0.9421)	(98, 0.9385)	(101, 0.9403)	(103, 0.9385)	(105, 0.9403)	(108, 0.9403)	(110, 0.9368)	(112, 0.9403)	(114, 0.9421)};
            \end{axis}
        \end{tikzpicture}
    }%
 \qquad
        \subfigure[QSAR]{
        \begin{tikzpicture}[remember picture,scale=0.55]
            \begin{axis}[
                    axis lines=left,
                    xmin=0,
                    xmax=220,
                    ytick={.64,.71,...,.87},
                    xtick={0,70,...,220},
                    ymin=.64,
                    ymax=.87,
                    thick,
                    width=0.66\textwidth,
                    height=0.34\textwidth
                ]
                \addplot[red,smooth,thick] plot coordinates {
                 (0, 0.6606)	(4, 0.6663)	(8, 0.6834)	(12, 0.6815)	(17, 0.708)	(21, 0.7231)	(25, 0.7355)	(29, 0.7344)	(34, 0.7629)	(38, 0.7866)	(42, 0.7838)	(46, 0.7856)	(51, 0.8065)	(55, 0.798)	(59, 0.8065)	(63, 0.8151)	(68, 0.8208)	(72, 0.8236)	(76, 0.8227)	(80, 0.8236)	(85, 0.8207)	(89, 0.8141)	(93, 0.8123)	(97, 0.8179)	(102, 0.8293)	(106, 0.8217)	(110, 0.8284)	(115, 0.8284)	(119, 0.8265)	(123, 0.8331)	(127, 0.8198)	(132, 0.8379)	(136, 0.8331)	(140, 0.8397)	(144, 0.835)	(149, 0.8349)	(153, 0.8359)	(157, 0.8369)	(161, 0.8416)	(166, 0.8368)	(170, 0.8397)	(174, 0.8426)	(178, 0.8445)	(183, 0.8473)	(187, 0.8445)	(191, 0.8445)	(195, 0.8493)	(200, 0.853)	(204, 0.855)	(208, 0.8511)	(212, 0.8558)};
                 \addplot[red,dashed,thick] plot coordinates {
                 (0, 0.6606)	(4, 0.691)	(8, 0.6767)	(12, 0.691)	(17, 0.7014)	(21, 0.7127)	(25, 0.7194)	(29, 0.7365)	(34, 0.7716)	(38, 0.7658)	(42, 0.7942)	(46, 0.7924)	(51, 0.8084)	(55, 0.7991)	(59, 0.7924)	(63, 0.7904)	(68, 0.7857)	(72, 0.7838)	(76, 0.7914)	(80, 0.799)	(85, 0.7933)	(89, 0.7857)	(93, 0.7905)	(97, 0.7829)	(102, 0.7895)	(106, 0.7924)	(110, 0.7942)	(115, 0.7999)	(119, 0.8019)	(123, 0.8009)	(127, 0.8009)	(132, 0.8047)	(136, 0.7962)	(140, 0.799)	(144, 0.8018)	(149, 0.818)	(153, 0.817)	(157, 0.8161)	(161, 0.8189)	(166, 0.8274)	(170, 0.8322)	(174, 0.8313)	(178, 0.8426)	(183, 0.8416)	(187, 0.8388)	(191, 0.8407)	(195, 0.8331)	(200, 0.8321)	(204, 0.8321)	(208, 0.8322)	(212, 0.836)};
                  \addplot[smooth,thick,dashed] plot coordinates {
                 (0, 0.6606)	(4, 0.6625)	(8, 0.6645)	(12, 0.7033)	(17, 0.7174)	(21, 0.7107)	(25, 0.7136)	(29, 0.7307)	(34, 0.7297)	(38, 0.7411)	(42, 0.7429)	(46, 0.7543)	(51, 0.7572)	(55, 0.7638)	(59, 0.7657)	(63, 0.7742)	(68, 0.7638)	(72, 0.7714)	(76, 0.7808)	(80, 0.7827)	(85, 0.7866)	(89, 0.7809)	(93, 0.7951)	(97, 0.798)	(102, 0.7998)	(106, 0.8055)	(110, 0.8027)	(115, 0.8056)	(119, 0.8027)	(123, 0.8055)	(127, 0.8046)	(132, 0.8046)	(136, 0.8055)	(140, 0.8122)	(144, 0.815)	(149, 0.8141)	(153, 0.815)	(157, 0.8169)	(161, 0.8131)	(166, 0.8112)	(170, 0.815)	(174, 0.816)	(178, 0.8169)	(183, 0.8198)	(187, 0.8179)	(191, 0.8179)	(195, 0.8179)	(200, 0.8141)	(204, 0.8198)	(208, 0.8217)	(212, 0.817)};
                  \addplot[smooth,thick,thick] plot coordinates {
                  (0, 0.6606)	(4, 0.6796)	(8, 0.7146)	(12, 0.71)	(17, 0.7545)	(21, 0.7554)	(25, 0.7754)	(29, 0.782)	(34, 0.7886)	(38, 0.8104)	(42, 0.8142)	(46, 0.8161)	(51, 0.8294)	(55, 0.819)	(59, 0.8067)	(63, 0.8237)	(68, 0.836)	(72, 0.8436)	(76, 0.8369)	(80, 0.837)	(85, 0.8455)	(89, 0.8436)	(93, 0.8483)	(97, 0.8436)	(102, 0.8407)	(106, 0.8436)	(110, 0.8464)	(115, 0.8474)	(119, 0.8474)	(123, 0.8435)	(127, 0.8483)	(132, 0.8436)	(136, 0.8446)	(140, 0.8531)	(144, 0.8588)	(149, 0.8512)	(153, 0.8455)	(157, 0.8492)	(161, 0.8578)	(166, 0.8559)	(170, 0.8634)	(174, 0.8653)	(178, 0.8616)	(183, 0.8587)	(187, 0.8606)	(191, 0.8616)	(195, 0.8578)	(200, 0.8578)	(204, 0.8587)	(208, 0.8492)	(212, 0.854)};
            \end{axis}
        \end{tikzpicture}
    }%
 \qquad
 \subfigure[methods]{
\begin{tikzpicture}[scale=0.5]
            \draw[red] (1,1)--(3,1);
            \node[right,red] at (4,1) {\small Epis};
            \draw[red, dashed] (7,1)--(9,1);
            \node[right,red] at (10,1) {\small Alea};
            \draw[black, dashed] (1,2)--(3,2);
            \node[right] at (4,2) {\small Rand};
            \draw[black] (7,2)--(9,2);
            \node[right] at (10,2) {\small Stan};
             \end{tikzpicture}}
           \caption{Average accuracies (y-axis) for logistic regression as a function of the number of examples queried from the pool (x-axis).}
           \label{fig:UncAccLogistic}
\end{figure}

For logistic regression, we start with a relatively small amount of initial training data, thereby making improvements in the beginning more visible. 
More specifically, the proportions of training/pool/test set are 1/89/10\,\%, and the accuracies are averaged. The budget is fixed to be $20\%$ of the original data, and in each   iteration, the learner is allowed to evaluate and query instances from a (randomly) subset consisting of $10\%$ data of the pool. 

\newpage 

In the case of logistic regression, the improvements through epistemic uncertainty sampling are less pronounced\,---\,on the contrary, the performance of epistemic and standard uncertainty sampling is quite comparable. Two examples, which are quite representative, are shown in Figure \ref{fig:UncAccLogistic}. As a plausible explanation, note that logistic regression comes with a very strong learning bias in the form of a linearity assumption. Therefore, the epistemic (or model) uncertainty disappears quite quickly.

\section{Conclusion} \label{sec:conlusion}

This paper reconsiders the principle of uncertainty sampling in active learning from the perspective of uncertainty modeling. More specifically, it starts from the supposition that, when it comes to the question of which instances to select from a pool of candidates, a learner's predictive uncertainty due to ``not knowing'' should be more relevant than its uncertainty due to inherent randomness. 

To corroborate this conjecture, we proposed \emph{epistemic uncertainty sampling}, in which standard uncertainty measures such as entropy are replaced by a novel measure of epistemic uncertainty. The latter is borrowed from a recent framework for uncertainty modeling, in which epistemic uncertainty is distinguished from aleatoric uncertainty \cite{senge:2014}. 
We interpret our experimental results, especially those for local learning (Parzen window classifier and decision trees) as evidence in favor of our conjecture. They clearly show that a separation of the total uncertainty (into epistemic and aleatoric) is effective, and that the epistemic part is the better criterion for selecting instances to be queried. This was the main purpose of the paper. 

Given this affirmation, we are now encouraged to elaborate on epistemic uncertainty sampling in more depth, and to develop it in more sophistication. This includes an extension to other learning algorithms and more general learning problems (such as multi-class classification), 
as well as a comparison to other variants of uncertainty sampling, such as \cite{antonucci:2012} and \cite{sharma:2017}.


\subsection*{Acknowledgements} This work was supported by the German Research Foundation (DFG) and the French National Agency for Research (Labex MS2T).


\end{document}